\documentclass[times,twocolumn,final,authoryear]{elsarticle}

\usepackage{ycviu}
\usepackage{framed,multirow}

\usepackage{amssymb}
\usepackage{latexsym}

\usepackage{url}
\usepackage{xcolor}
\definecolor{newcolor}{rgb}{.8,.349,.1}

\usepackage{comment}
\usepackage{amsmath} 
\usepackage{eso-pic,xspace}
\usepackage{color}
\usepackage{floatrow}
\usepackage{array}
\usepackage{colortbl}	
\usepackage{epsfig}
\usepackage{graphicx}
\usepackage{tabularx,ragged2e}
\usepackage[export]{adjustbox}
\usepackage{booktabs,calc}
\usepackage{hhline}
\usepackage{float} 
\usepackage{subfig}
\usepackage[scientific-notation=true]{siunitx}
\usepackage{type1cm}
\usepackage{lettrine}
\usepackage{rotating} 
\usepackage{longtable}
\usepackage{wrapfig}
\usepackage{stmaryrd} 
\usepackage{setspace}
\usepackage{cuted}
\usepackage{capt-of}
\usepackage{xspace} 
\usepackage{hyperref}
\usepackage{booktabs}
\usepackage{scrextend} 
\hypersetup{
 colorlinks=true,
 linkcolor=red,
 citecolor=green,
}

\makeatletter
\DeclareTextFontCommand{\emph}{\em} 
\DeclareRobustCommand\onedot{\futurelet\@let@token\@onedot}
\def\@onedot{\ifx\@let@token.\else.\null\fi\xspace}
\def\eg{e.g\onedot} 
\def\ie{i.e\onedot}

\journal{Computer Vision and Image Understanding}

\begin{document}

\begin{frontmatter}

\title{Transformed ROIs for Capturing Visual Transformations in Videos}

\author[1]{Abhinav \snm{Rai}} 
\author[2]{Fadime \snm{Sener}\corref{cor1}}
\cortext[cor1]{Corresponding author: 
 }
\ead{sener@cs.uni-bonn.de}
\author[1]{Angela \snm{Yao}}

\address[1]{School of Computing, National University of Singapore, Singapore 117417, Republic of Singapore}
\address[2]{Institute of Computer Science II, University of Bonn, 53115 Bonn, Germany}

\begin{abstract}
Modeling the visual changes that an action brings to a scene is critical for video understanding. Currently, CNNs process one local neighbourhood at a time, thus contextual relationships over longer ranges, while still learnable, are indirect. We present TROI, a plug-and-play module for CNNs to reason between mid-level feature representations that are otherwise separated in space and time. The module relates localized visual entities such as hands and interacting objects and transforms their corresponding regions of interest directly in the feature maps of convolutional layers. With TROI, we achieve state-of-the-art action recognition results on the large-scale datasets Something-Something-V2 and EPIC-Kitchens-100. 
\end{abstract}

\end{frontmatter}

\section{Introduction}

In this paper, we target the task of recognizing human-object interactions found in daily activities. These actions are fine-grained and inherently compositional between the movement (verb) and the interacting object. Throughout the interaction, the objects are often transformed visually and/or physically, and over longer periods of time. Correctly identifying these actions requires the ability to relate the object's appearance from beginning to end throughout the transformation. 
State-of-the-art convolutional neural networks (CNNs), armed with 3D convolutions, are designed to learn such relationships in space and time. However, convolution operations are, by design, locally limited and are therefore inefficient in capturing relationships over a long range. 

One way to expand a network's receptive field over time and space would be to increase the network depth. However, adding more layers na\"ively also increases the memory requirements and computations~\citep{he2015convolutional} and runs the risk of over-fitting. Instead, architectural additions in the form of skip connections~\citep{huang2017densely}, attention~\citep{wang2018non,chenA2} and graphs~\citep{chen2019graph} have been proposed, all with the aim of capturing either long-range or global dependencies more directly.

\begin{figure}[t]
\centering 
\includegraphics[width=0.8\linewidth,trim=3cm 0cm 7.2cm 1.2cm,clip]{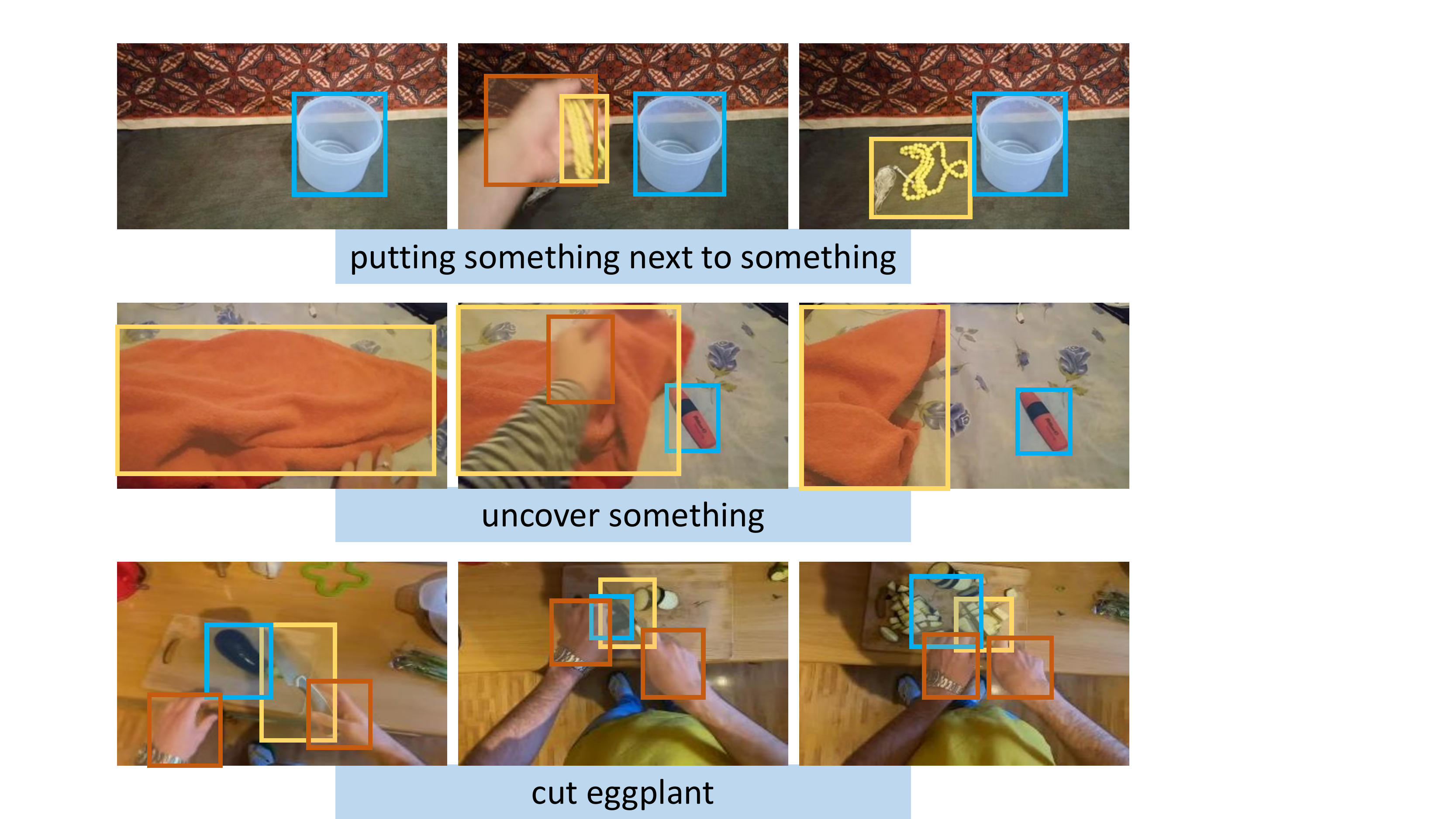}
\caption{
Our model captures long-range and/or global transformations of visual entities to learn hand-object interactions.} 
\label{fig:transformations}
\end{figure}

With the recent advances in object detectors and trackers, a commonly adopted strategy is to first localize key entities such as body parts and/or interacting objects. The action is then recognized based on the detections or tracks, with additional reasoning through graphs~\citep{wang2018videos}, visual relations~\citep{baradel2018object}, or other forms of interaction networks~\citep{materzynska2020something}. To represent the detections and tracks, features are drawn from the output of the CNN's last convolutional layer~\citep{wang2018videos,baradel2018object}.

Purely feature-based and purely entity-based approaches represent two extremes in learning global dependencies in video. On the one hand, feature-based approaches like non-local blocks and GloRe~\citep{wang2018non,chen2019graph} exhaustively relate locations on feature maps. This results in a combinatorially large number of operations, even though the majority of these computations are redundant as most of the locations make little contribution to others. Furthermore, because these methods capture global relationships from the entire feature map indiscriminately, they cannot ensure that learned relationships correspond to causal entities\footnote{``Causal'' here refers to the fact that the action classes in Something-Something and EPIC-Kitchens are defined by the composition of a verb (movement of the hands) and nouns (interacting objects).}, such as the hands and interacting objects. This, in turn, may lead to undesirable biases to the scene~\citep{li2018resound,choi2019can} or other dataset latencies. On the other hand, purely entity-based approaches reason only between detections or tracks. They miss useful cues from the background and lack the ability to reason between entities and the scene itself. Therefore, to boost performance, most approaches~\citep{baradel2018object,wang2018videos,materzynska2020something} add a separate appearance stream and perform a late fusion to merge this information.
 
\begin{figure*}[!htb]
\centering 
\includegraphics[width=0.89\linewidth,trim=0cm 6.4cm 5.5cm 0cm,clip]{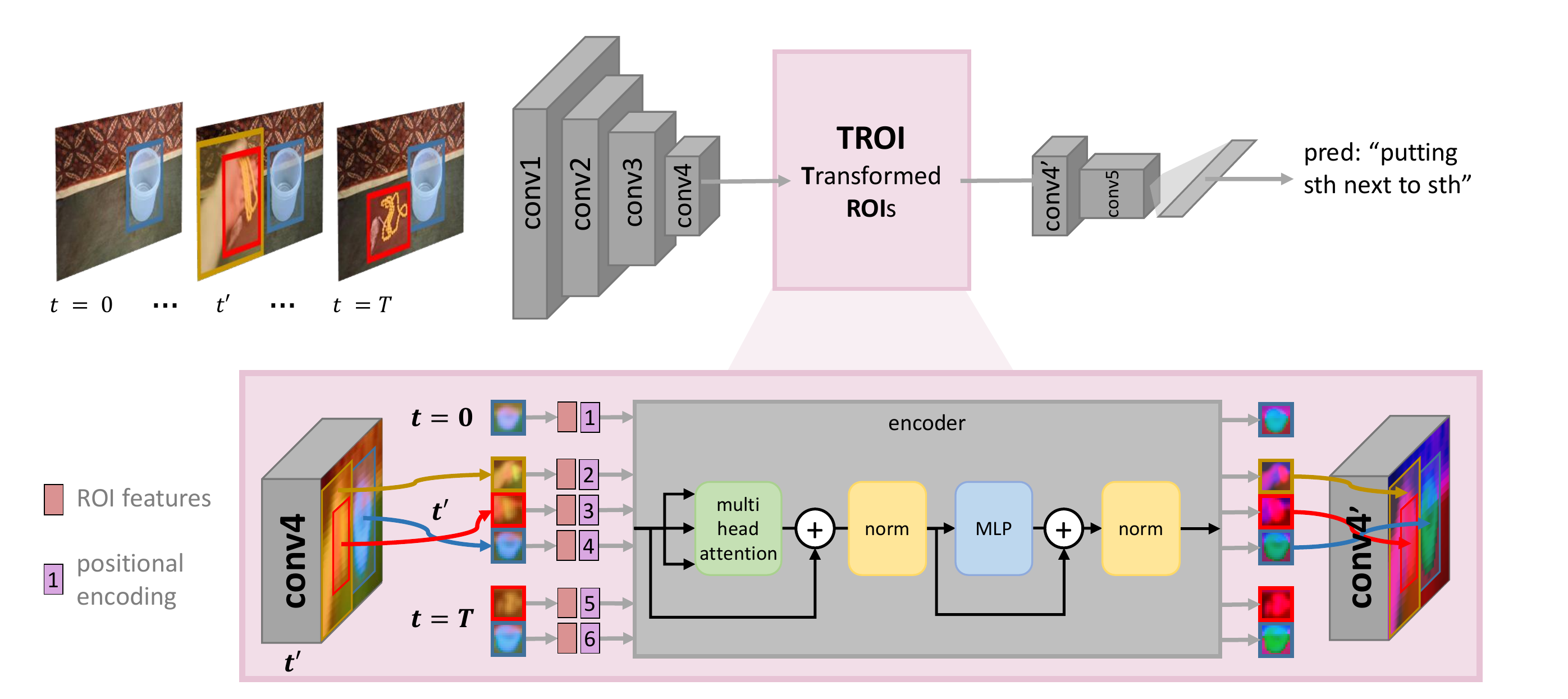}
\caption{\textbf{Overview of TROI.} The module can be integrated between the convolution layers of standard CNNs; above, TROI is inserted after the \emph{conv4} layer. Within TROI, a RoIAlign operation localizes the features for each visual entity, and a transformer-style encoder relates these mid-level representations in space and time. To retain spatial and temporal information from the input ROIs, we utilize positional encoding. The transformed features from hand and object regions are updated in place on the feature maps. The updated feature maps, \ie \textbf{conv4'} above, are followed by the last convolutional layer and a classification layer. 
} 
\label{fig:intro_overviwe}
\end{figure*}

We aim to strike a middle ground between abstract visual features and concrete visual entities. Relying solely on high-level visual entities may not allow us to capture the dynamic changes in objects' appearance; lower-level visual features can capture these differences, but at the same time may lack sufficient semantic context. Therefore, we propose a module for regions of interest (ROIs) associated with visual entities on mid-level feature maps. Representations of the ROIs are transformed in-place, directly within the feature map. Accordingly, we name our proposed module TROI (\textbf{T}ransformed \textbf{ROI}s). Based on detections, the module associates ROI features across an entire video via an attention mechanism. This localized form of attention bypasses the locality constraints of CNN backbones and allows the ROIs to be related to each other in a manner that is global and/or long-range in time. The transformed ROIs continue to be processed within their feature map context by the original CNN backbone to capture global information about the entire video content. An overview is presented in Figure~\ref{fig:intro_overviwe}. 

The TROI module allows us to benefit from localizing visual entities through detectors and trackers while not being entirely limited to high-level semantics, which may absorb the finer-grained appearance differences. For example, in the bottom of Figure~\ref{fig:transformations}, the whole and half-sliced eggplants have very different appearances due to the cutting action. At the final layers of a CNN, they may both be correctly identified as \emph{`eggplant'}, or the sliced eggplant may be confused with other (more commonly sliced) objects. Either way, it is highly challenging to model the fine-grained transformations of the appearance. Similarly, for \emph{`uncovering something'} in the middle of Figure~\ref{fig:transformations}, the pen is revealed by the hand at some point during the sequence. If the transition is not captured, the sequence could resemble the action of \emph{``putting something next to something''} shown in the top of Figure~\ref{fig:transformations}. TROI allows CNNs to capture such transformations implicitly while using explicit positional information. 

Our work is novel in several regards. First and foremost, our method is the first to focus on relating mid-level features over space and time in middle CNN layers from an entity point of view. This allows us to bypass the large number of operations on irrelevant regions, contrary to feature-based approaches~\citep{wang2018non,chen2019graph}. Our in-place transformations in the ROIs unify entity localization and action recognition in CNNs, contrary to other approaches that separate the two in a CNN’s last convolutional layer. Finally, our approach is flexible. Any state-of-the-art relational model can be used to represent spatio-temporal inputs; we choose transformers but also demonstrate the feasibility with GCNs. We summarize our contributions below:

\begin{itemize}

\item We propose a new approach for reasoning on mid-level feature maps that captures relationships between detected entities as they transform over time. 

\item We present a flexible and lightweight module, TROI, that can be integrated into standard CNNs, such as 3D ResNet~\citep{carreira2017quo} and TSM~\citep{lin2019tsm}, to transform regions directly at the feature-map level. TROI uses self-attention to facilitate the global modelling of visual entities and encodes the transformation in place. 

\item The proposed module achieves the state-of-the-art on two large-scale fine-grained action recognition datasets. Our improvements on Something-Something-V2~\citep{goyal2017something} and EPIC-Kitchens-100~\citep{damen2018scaling} over the state-of-the-art highlight the effectiveness of our strategy in relating mid-level features. 

\end{itemize}

\section{Related Works}

\textbf{Convolutional neural networks} are the dominant approach for action recognition from video. State-of-the-art methods are 3D CNNs~\citep{tran2015learning,carreira2017quo,feichtenhofer2019slowfast,xiao2020audiovisual} that extend convolution and pooling to the time dimension, although they tend to be large and expensive models. To balance speed and accuracy, some layers either reduce 3D convolutions into 2D~\citep{xie2018rethinking} or factorize them into 2D spatial and 1D temporal convolutions~\citep{tran2018closer}. Others have found channel shifting~\citep{lin2019tsm,fanbuch2020rubiks} to be highly effective at boosting discriminativeness while limiting computational expense. Our work is orthogonal to these methods in that we propose an additional module that can be integrated into any of the aforementioned architectures, and we demonstrate this on a 3D ResNet~\citep{carreira2017quo} and TSM~\citep{lin2019tsm}.

\textbf{Relational reasoning} aims to directly capture interactions between human-centered entities such as people, objects, or hands. Recent work either reasons implicitly between visual features or explicitly between localized entities. The former has been applied to visual Q\&A~\citep{santoro2017simple}, action localization~\citep{sun2018actor}, and temporal interactions~\citep{zhou2018temporal}. \cite{wang2018non} proposes adding a non-local layer to 3D CNNs, which correlates features from all locations using self-attention. Follow-up works have tried to reduce the computational cost of such exhaustive pixel-level relations~\citep{chenA2,zhang2019latentgnn,chen2019graph}. $\text{A}^2$Net~\citep{chenA2} generates a set of global features to propagate the information between positions more efficiently. GloRe~\citep{chen2019graph} learns a latent interaction space and uses a fully connected graph to extract relation-aware features. Note that these approaches are region-agnostic and use the entire feature map indiscriminately. As such, they cannot guarantee correspondences to specific entities; furthermore, they perform a large number of operations on irrelevant regions.

Localizing ROIs to use explicit reasoning is well-suited for scenes that are cluttered an/or uninformative for identifying actions. This is often the case in egocentric actions where hands are in contact with the active objects~\citep{pirsiavash2012detecting,damen2018scaling}. 
\cite{wang2018videos} construct spatio-temporal graphs over object proposals. Similarly, the interaction network from \cite{materzynska2020something} models spatio-temporal geometric relations between objects in object-graph representations. 
The Tube Convolutional Neural Network (T-CNN)~\citep{hou2017tube} introduces tube of interest (ToI) pooling in the latter CNN layers for spatio-temporal action localization.
\cite{baradel2018object} model the inter-frame interactions between object segments using pairwise mappings and an RNN. 
Recent works by ~\cite{wang2020symbiotic} and \cite{wang2020symbioticTpami} compute object-centric features via an attention mechanism that encourages mutual interactions between object regions and enhances action-relevant interactions.

We position our method as being closer to the second group as we conduct relational reasoning over pre-computed region proposals. However, our approach does not need to model the global appearance of frames in a separate stream to support the recognition performance. 
A very recent work by \cite{herzig2021object} performs an in-place feature update similar to our work, but using transformers as a backbone.

\textbf{Relating Entities:}
Graphs are a natural way to relate visual entities, and graph convolutional networks~\citep{kipf2016semi} are popular for relational reasoning in action recognition~\citep{wang2018videos,chen2019graph}. However, previous works are inflexible in that they assume a fixed graph structure. Within the graph, the temporal ordering between the nodes is lost~\citep{chen2019graph}. Additionally, as the graphs are often fully connected over entire sets of features, the number of input nodes must be limited~\citep{wang2018videos}. In both cases, the data structure cannot accommodate variations of real sequences, \eg when the number of relevant entities changes over time or is simply not detected. TROI aims to fill these two shortcomings by explicitly augmenting the entity representations with positional encodings, and it also allows for a variable number of input entities.

Finally, we note that our work may bear similarity to the Video Action Transformer Network of~\cite{girdhar2019video} as we both use transformers. However, we have several key differences in terms of the objective and architecture. The Transformer Network uses action recognition as a supervisory cue to learn transformers for localizing and tracking actions. Our module, on the other hand, uses already localized ROIs to improve recognition. Architecture-wise, ~\cite{girdhar2019video}'s model includes three times more parameters than TROI and is trained over a full week using 10 GPUs. In contrast, TROI is a lightweight module that can be integrated into many CNNs to transform regions directly at the feature-map level. It adds a negligible parameter count on top of CNN backbones and trains with a order of magnitude less in terms of both time and hardware.

\section{Method}

\subsection{Preliminaries}
Given a video, we select $T$ evenly interspersed frames, as is standard in CNNs for action recognition~\citep{lin2019tsm,feichtenhofer2019slowfast}. Let $\textbf{X} \in \mathbb{R}^{T \times W \times H \times C}$ denote a feature representation of the video, where $W$, $H$ and $C$ are the width, height, and number of channels of the feature map, respectively. $\mathbf{X}$ can come from any convolutional layer in a neural network; as we are interested in relating mid-level features, $\mathbf{X}$ is the feature map after several convolution operations, 
\eg at conv3 or conv4 of a ResNet architecture.

On a selected frame $t$, there are $N_t$ regions of interest (ROIs) around fixed entities such as the hands or interacting objects. We denote the collection of ROIs across the $T$ selected frames as $\mathcal{R} = \{r_1, ..., r_N\}$, with $N$ ROIs in total. The objective of our approach is to learn a mapping $g$, parameterized by $\Theta$, to map $\textbf{X}$ to $\textbf{X}'$, \ie
\begin{equation}
 \textbf{X}' = g(\textbf{X}, \mathcal{R}, \Theta).
\end{equation}

\noindent $\textbf{X}'$ has the same dimensions as $\textbf{X}$ but is transformed in that the $N$ ROI regions are modified according to the learned relationships between all the entities in the video. To do this, we propose a transformer-based self-attention module, which we name TROI~(\textbf{T}ransformed \textbf{ROI}s).

We denote with $f_i \in \mathbb{R}^{C}$ the visual feature within the ROI $r_i$, where $f_i = h(X_i,r_i)$. Here, $h$ is a function that localizes the features corresponding to ROI $r_i$, and $X_i \in \mathbb{R}^{1 \times W \times H \times C}$ is the part of the feature map selected from the frame to which $r_i$ corresponds. In practice, we estimate the ROI $r_i$ based on object detections and localize the associated feature from $X_i$ by applying ROIalign~\citep{he2017mask} followed by a spatial average pooling for $h$. To retain positional information from the bounding boxes in space and time, we add a positional encoding~\citep{vaswani2017attention} to each $f_i$. For the positional encoding, we follow a left-to-right order for the bounding boxes across the frames by labelling them sequentially, as shown in Figure~\ref{fig:intro_overviwe}. We found similar results when ordering from right to left, which indicates that TROI consistently learns patterns as long as some temporal and spatial ordering is preserved. We then use this as the positional value and employ the sine and cosine formula used by \cite{vaswani2017attention} to generate the final encoding. The entire set of features associated with all the ROIs extracted from the video's feature maps $\mathbf{X}$ is then denoted as $\mathbf{F} \in \mathbb{R}^{N \times C}$.

\subsection{Relational Module}

For the mapping $g$ and its parameters $\Theta$, we choose a transformer-style encoding architecture and leverage self-attention to learn the spatio-temporal relationships between the visual entities. As an alternative to $g$, one can also use, \eg, a GCN (see ablations in Section~\ref{lab:module_ablation}), but we opt for a transformer due to its stronger performance. 

In video, transformers~\citep{vaswani2017attention} have been used for video encoding~\citep{sun2019videobert}, action localization~\citep{girdhar2019video}, and video captioning~\citep{zhou2018end}. Transformers compute sequence representations by attending to different input positions via stacks of self-attention layers. In this context, we adapt them to model the set of mid-level ROI features in a video as a natural spatio-temporal sequence. 

We follow standard transformer notation and use $\mathbf{qkv}$ self-attention~\citep{vaswani2017attention}, with query $\mathbf{q}$, key $\mathbf{k}$ and value $\mathbf{v}$. Conceptually, we perform a series of in-place transformations dependent on the other ROIs from the entire video through self-attention. A \textbf{query} is an input ROI feature, and the \textbf{keys} are the sequence of features from all the other ROIs in a video. The self-attention mechanism estimates attention weights that correspond to the importance of the ROIs keys for a particular query ROI. According to this, we first compute the linear projections for $\mathbf{q}$, $\mathbf{k}$ and $\mathbf{v}$ features :

\begin{equation}
 [\mathbf{q},\mathbf{k}, \mathbf{v}] = \mathbf{F} \mathbf{W}_{qkv},
\end{equation}

where $\mathbf{W}_{qkv} \in \mathbb{R}^{C \times (2d_k+d_v)} $ corresponds to the parameter matrices $\mathbf{W}_q \in \mathbb{R}^{C \times d_k}, \mathbf{W}_k \in \mathbb{R}^{C \times d_k} , \mathbf{W}_v \in \mathbb{R}^{C \times d_v} $ of the projected versions of query, key and value, respectively. Here, $d_k = d_v = C/m$ and $m$ is the number of attention heads. We refer the reader to the work of~\cite{vaswani2017attention} for more details on the transformer architecture.

We compute the attention weights, $A_{ij}$, as scaled pairwise similarities between two ROI features from $\mathbf{F}$ and their query and key representations, $\mathbf{q}^i$ and $\mathbf{k}^j$,

\begin{equation}
 \mathbf{A} = \text{softmax}\left(\frac{\mathbf{q}\mathbf{k}^T}{\sqrt{d_k}}\right), \quad\quad \mathbf{A} \in \mathbb{R}^{N \times N}.
\end{equation}

\noindent For computing the self-attention, SA, for each ROI in $\mathbf{F} \in \mathbb{R}^{N \times C}$, we compute a weighted sum over all \emph{values} $\mathbf{v}$, \ie

\begin{equation}
 \text{SA}(\mathbf{F}) = \mathbf{A}\mathbf{v}.
\end{equation}

The self-attention operation calculates the response at each position, $f_i$, by attending over all other positions in $\mathbf{F}$. Multi-head attention extends self-attention by running the same procedure $m$ times; each self-attention procedure is referred to as a \emph{head}. Multi-head attention has an implicit ensembling effect and generally improves performance by concatenating multiple self-attention outputs:

\begin{equation}
 \text{MHA}(\mathbf{F}) = 
 [\text{SA}_1(\mathbf{F}), \text{SA}_2(\mathbf{F}), \ldots , \text{SA}_m(\mathbf{F})] \mathbf{W}_{m},
\end{equation}
where the query, value and key projections are found in the parameter matrix $\mathbf{W}_{m} \in \mathbb{R}^{m \cdot d_v \times C}$.

Let $\ell$ be the index of one encoder layer. Then, $f'_{i,\ell} \in \mathbb{R}^{C}$ denotes visual information updated with self attention, and $\mathbf{F}'_\ell \in \mathbb{R}^{N \times C}$ the entire set of updated features in layer $\ell$. We compute these features as
\begin{equation}
\begin{array}{l}
\mathbf{G}'_\ell = \text{LN}\left(\text{MHA}(\mathbf{F'}_{\ell-1}) + \mathbf{F}'_{\ell-1}\right), \\ 
\mathbf{F}'_\ell = \text{LN}\left(\text{MLP}(\mathbf{G}'_\ell) + \mathbf{G}'_\ell\right) \\ 
\end{array}
\end{equation}
where $\mathbf{F}'_\ell$ corresponds to the updated region features after every encoder layer. The updated features from the previous layers, $\mathbf{F}'_{\ell-1}$, are used as input in the following layers. The multi-layer perceptron (MLP) contains two linear transformations with ReLU activations in between. LN corresponds to layer normalization~\citep{ba2016layer}. We update the original feature map $\mathbf{X}$ with the updated features, $\mathbf{F}'$ with each $f_i$ in place, so that the relative spatial positioning of the features from $h$ is retained in the transformed $\mathbf{X}'$. 
The resulting in $f' \in R^{1\times 1\times C}$. We then replicate $f'$, $W\times H$ times, to expand it to a size of $W \times H \times C$. Depending on the feature level of $\textbf{X}$ and $\textbf{X'}$, $W$ and $H$ are very small, are either $1$ or $2$. As such, the replication does not introduce significant redundancy.
As a result, we only modify the features for the ROI regions; the rest remain unchanged.

The relational module in TROI was designed with similar motivations as other relational frameworks, such as non-local networks~\citep{wang2018non} and GloRe~\citep{chen2019graph}. We share some basic components, like self-attention, however, we differ in a key aspect in that non-local networks and GloRe compute the relations between every feature position and all others on the feature maps, $\mathbf{X}$, exhaustively. They do not leverage any explicit entity-aware localization, $\mathbf{F}$, so they perform many more redundant relations and computations.

\subsection{Implementation Details}

\textbf{Training:}
Our models are trained for 80 epochs with a batch-size of 64 over two 48GB GPUs. We use SGD optimization with a momentum of 0.9 and a weight decay of 5e-4. The learning rate is set at 0.01 for the initial 20 epochs, is decreased to 0.001 for the following 20 epochs, and is finally set to 10e-4 for the remaining epochs. The backbone network is initialized with pre-trained Kinetics~\citep{carreira2017quo} weights. 
We then add TROI and train using the strategy outlined above.
For the transformer, two attention heads are used. 
We use a single classification head to directly predict the action class.

To sample images from the video, we follow~\cite{wang2016temporal} and ~\cite{lin2019tsm}. A fixed number of frames (8 or 16) is selected evenly in time from each video clip. Using more frames, \ie, 16 instead of 8, typically improves the performance at the expense of larger models. Unless otherwise indicated, we use 8 frames in our experiments.
Note that our approach considers each frame individually without temporal association across frames and does not require the class of objects to be known.
For data augmentation, we use the strategy suggested by~\cite{lin2019tsm}: scale jittering, corner cropping, and random horizontal flipping. The augmented image is then resized to $224\times 224$. 
 
\textbf{Inference:} 
We follow the same sampling procedure and data-preprocessing strategy as~\cite{lin2019tsm} for EPIC-Kitchens-100 and~\cite{goyal2017something} for Something-Something-V2 to ensure a fair comparison with other works.

\section{Experiments}
 
\subsection{Datasets \& Evaluation} 
 
We experiment on two large-scale action datasets:
 
\noindent \textbf{Something-Something-V2} has 220K trimmed videos with 174 fine-grained action classes of human-object interactions, such as \emph{``plugging something into something''}. We use two types of bounding boxes for training, both provided by~\cite{materzynska2020something}: The first is based on ground truth annotations~\footnote{7\% of training and 9\% of validation videos do not have annotations; we simply bypass our relation module for these videos.}, and the second is based on the per-frame bounding box predictions from a multi-object tracking approach~\citep{bewley2016simple} applied to the outputs of an object detector~\citep{wu2019detectron2} for the hand and interacting objects. We evaluate on both prediction outputs and GT annotations for a fair comparison with~\cite{materzynska2020something}. We report the Top-1 and Top-5 accuracies on the validation set using RGB images. Following the methodology of~\cite{lin2019tsm} and \cite{fanbuch2020rubiks}, we sample either ``1-Clip'' with one center crop of $224\times224$, or ``2-Clips'' with three crops (left, right and center) of $224\times224$. Our network is trained with 8 or 16 frames as inputs.

\textbf{EPIC-Kitchens-100} features egocentric recordings of 37 subjects. There are around $90$K pre-trimmed segments featuring fine-grained actions composed of verb and noun classes, \eg, \emph{``pour water''}. The dataset contains 4,025 actions composed of 97 verbs and 300 nouns. It provides RGB and optical flow images, as well as bounding boxes extracted by a hand-object detection framework~\citep{shan2020understanding} that returns class labels left/right hand and left/right interacting object. For each frame, there are up to four bounding boxes for these classes. On average, 70\% and 76\% of the frames have left and right hand/object detections, respectively. For EPIC-Kitchens, we train and test our methods with offline detections from ~\cite{shan2020understanding}.

Following the EPIC-Kitchens protocol of ~\cite{damen2020rescaling}, we report Top-1/5 accuracies for both the validation and test sets. The validation and test sets are further split to distinguish \textit{unseen} participants and \textit{tail} classes to measure the model generalization. Tail classes are defined as those contributing to less than 20\% of the total instances in the training set. However, as EPIC-Kitchens is a very long-tailed dataset, 88\% of the action classes\footnote{3535 actions, 86 verbs, 228 nouns} are considered tail classes, making it highly challenging.

\subsection{Impact of Adding the TROI Module}
In Table \ref{tab:effect_modality_bck}, we first compare the performance with and without our proposed module on the validation set of EPIC-Kitchens for two state-of-the-art action recognition backbones: 3D ResNet-50 and TSM. We train separate models for RGB and optical flow for predicting the verbs and nouns and average pre-softmax predictions during inference when fusing the two models. In all comparisons, we use a single-layer TROI inserted at conv4, as verified by ablation studies in Sec.~\ref{lab:module_ablation}.

On both backbones, integrating our proposed module improves the verb and noun performance by approximately 1\% for all the modalities. On the 3D ResNet backbone, we have gains of 1.3\% and 0.9\% verb accuracy and 1.6\% and 1.4\% noun accuracy for RGB and flow, respectively. On the TSM backbone, the gains are smaller for RGB (+0.8\% for verb +0.6\% for noun). When fusing the two modalities, adding TROI gains 1.5\% of verb and 1.7\% for noun accuracy for 3D ResNet and 0.5\% verb \& 1.2\% noun accuracy for TSM. We speculate that TSM has smaller gains as it is already a very strong baseline; it outperforms 3D ResNet-50 by 8.1\% using the two modalities alone for verb accuracy and 3.8\% for noun accuracy.

Could the improvements come simply from localizing the visual entities, \ie the hands and objects? We verify that this is not the case in Table~\ref{tab:importance_bck}. The first baseline, TSM, has no location information on the visual entities and simply performs action recognition using the entire frame. The second baseline, ROI, ignores all information outside the bounding boxes and classifies only what is within them. Similar to~\cite{girdhar2019video}, this experiment corresponds to directly taking ROI features and applying a fully connected layer after the transformer-encoder for classification. We can see that the in-place transformation of the regions in the mid-level layers of the CNN is beneficial and outperforms~\cite{girdhar2019video}. We observe a drop of around 14\% and 16\% in action accuracy compared to TSM when using ROI in EPIC-Kitchens and Something-Something, respectively. 
TSM + ROI fuses the two predictions; the accuracies are close to the TSM baseline. TROI, however, results in the best performance and outperforms the TSM baseline by 0.8\%, 0.6\% and 0.4\% for verb, noun and action Top-1s, respectively, on EPIC-Kitchens and by 3.8\% on Something-Something.

\begin{table}[htb!] 
\centering
\resizebox{\linewidth}{!}{
\setlength{\tabcolsep}{4.1pt}
\begin{tabular}{@{}lcccll@{}}
\toprule
 & & \multicolumn{2}{c}{Verb Top-1} & \multicolumn{2}{c}{ Noun Top-1 } \\ \cmidrule(l){3-6} 
Backbone & Data & w/out TROI & w/ TROI & \multicolumn{1}{c}{ w/out TROI} & \multicolumn{1}{c}{ w/ TROI } \\ \midrule
3D ResNet & RGB & 55.95 & 57.26 & 42.58 & 44.17 \\
3D ResNet & Flow & 53.52 & 54.44 & 33.40 & 34.83 \\
3D ResNet & RGB + Flow & 60.28 & 61.75 & 45.11 & 46.88 \\ \midrule
TSM & RGB & 62.91 & 63.71 & 46.88 & 47.50 \\
TSM & Flow & 63.84 & 64.84 & 35.84 & 35.85 \\
TSM & RGB + Flow & 68.35 & 68.80 & 48.99 & 49.22 \\ \bottomrule
\end{tabular}
}
\caption{Comparison of different backbones and modalities over Top-1 verb and noun accuracy} on the validation set of EPIC-Kitchens. 
\label{tab:effect_modality_bck} 
\end{table}

\begin{table}[!htb]
\centering
\resizebox{\linewidth}{!}{
\setlength{\tabcolsep}{4.0pt}
\begin{tabular}{@{}ccccc@{}}
\toprule
 & \multicolumn{3}{c}{EPIC } & Something \\ \cmidrule(l){2-5} 
\multicolumn{1}{c|}{Method} & Verb & Noun & \multicolumn{1}{c|}{Action} & Action \\ \midrule
\multicolumn{1}{l|}{TSM} & 62.91 & 46.88 & \multicolumn{1}{c|}{35.45} & 58.82 \\
\multicolumn{1}{l|}{ROIs~\citep{girdhar2019video}} & 49.62 & 30.29 & \multicolumn{1}{c|}{21.30} & 47.94 \\
\multicolumn{1}{l|}{TSM + ROIs~\citep{girdhar2019video}} & 62.27 & 46.14 & \multicolumn{1}{c|}{34.28} & 59.22 \\
\multicolumn{1}{l|}{TROI} & 63.71 & 47.50 & \multicolumn{1}{c|}{36.23} & 64.70 \\
\multicolumn{1}{l|}{TSM+TROI} & 64.73 &49.15 & \multicolumn{1}{c|}{37.66} & -\\ 
\bottomrule
\end{tabular}
}
\caption{Effect of localization. 
We use GT annotations for Something-Something. ROIs and TROI use a TSM backbone. Reported are Top-1 accuracies on the validation set for RGB images.}
\label{tab:importance_bck} 
\end{table}

\subsection{Module Ablation Studies}\label{lab:module_ablation}

Below, we validate our design choices on EPIC-Kitchens-100 and Something-Something's validation sets for RGB images. 

\textbf{Module layers and placement:} We first assess the number of encoder layers and the module's placement relative in the backbone in Table~\ref{tab:ablation_transformer}. We achieve our best performance with a single encoder layer; the performance is comparable with two layers. Regarding the placement for our module, we find that a location after higher convolutional layers, like conv4 and conv5, provides better results than conv3. This is likely because these layers contain higher-level semantics relevant to the ROIs. If TROI is placed after conv5, the performance degrades slightly; this is likely due to the low resolutions of the feature maps in the later layer, \ie, $7\!\times\!7$, in which small object regions are no longer distinguishable. We also insert multiple blocks on different layers and find that adding TROI in conv4 and conv5 gives the best results; however, the verb recognition performance of this ablation is 1.9\% lower and the noun performance is 3.1\% lower than a single TROI in conv4. These observations align with~\cite{chen2019graph}, who also report conv4 as the optimal location for their relational method. 

\textbf{Transformer vs. GCN:} We next assess the choice of architecture for relational learning and compare the use of a transformer versus a graph convolutional network (GCN). GCNs are the most popular choice in learning relations and are used by ~\cite{chen2019graph} and ~\cite{wang2018videos}. Similar to ~\cite{chen2019graph}, we use a single single-layer graph convolution network. Our findings in Table~\ref{tab:gcn_vs_transformer} show that using a GCN does not improve upon the original backbone, while TROI increases the verb accuracy by 1.3\%. We believe that the superior performance comes from our explicit temporal modeling via positional encoding, whereas the GCN treats all ROIs as nodes of a fully connected graph. We further evaluate GCNs by using GloRe~\citep{chen2019graph}, which is a recent relational method that uses graph convolutions to reason (exhaustively) over individual features. GloRe gives 0.4\% verb improvement over a 3D ResNet-50 backbone, but Top1 verb accuracy of TROI is still 0.8\% better than GloRe while having fewer parameters. This indicates that integrating localized reasoning, \ie, on the object and hand regions, is more effective than purely feature-based approaches.

\begin{table}[!htb]
\centering
\resizebox{\linewidth}{!}{
\setlength{\tabcolsep}{15.1pt}
\begin{tabular}{@{}lllll@{}}
\toprule
Model & \#Layers & Block & Verb Top1 & Noun Top1 \\ \midrule
TROI & 1 & conv4 & 57.26 & 44.17 \\ \midrule
TROI & 2 & conv3 & 54.52 & 36.09 \\
TROI & 2 & conv4 & 57.04 & 44.34 \\
TROI & 2 & conv5 & 56.63 & 41.30 \\ \bottomrule
\end{tabular}
}
\caption{Ablations on design choices of the TROI module reported on the validation set of EPIC-Kitchens for RGB inputs. Reported are Top1 verb and noun accuracies.
}
\label{tab:ablation_transformer} 
\end{table}

\begin{table}[!htb]
\centering
\resizebox{\linewidth}{!}{
\setlength{\tabcolsep}{12.7pt}
\begin{tabular}{@{}lllll@{}}
\toprule
Model&3D ResNet-50 & GCN &GloRe & TROI \\ \midrule
Verb Top-1 & 55.95 & 55.89 & 56.37 & \textbf{57.26} \\ 
 Noun Top-1 & 42.58 & - & - & 44.17 \\ \hline 
\# Params (M) & 28.42 & 29.46 & 34.17 & 33.67 \\ \hline
\end{tabular}}
\caption{Comparison of TROI to a GCN and GCN-based relational module GloRe~\citep{chen2019graph}. A 3D ResNet-50 backbone is used for all methods; the results are reported on the validation set of EPIC-Kitchens for RGB trained for predicting the verbs and nouns. 
Note that our model requires hand-object bounding boxes during inference. Given that the backbone model is shared, a SOTA object detector~\citep{ren2015faster} would add 19M additional parameters. 
}
\label{tab:gcn_vs_transformer} 
\end{table}

\begin{table}[!htb]
\centering
\resizebox{\linewidth}{!}{
\setlength{\tabcolsep}{17.7pt}
\begin{tabular}{@{}lll@{}}
\toprule
Visual Entities & Something Top1 & EPIC Top1 \\ \midrule
All hands \& objects & 62.60 & 36.40 \\\midrule
Only objects & 61.78 & 35.83 \\
Only hands & 56.10 & 33.41 \\
Right hand \& object & - & 35.18 \\
Left hand \& object &- & 34.61 \\\midrule
No entities & 54.25 & 32.04\\\midrule
All entities, IOU@0.05 & 54.59 & 30.45 \\
All entities, IOU@0.25 & 59.12& 34.97 \\
All entities, IOU@0.50 & 61.75& 35.73 \\
\midrule
\end{tabular}}
\caption{Influence of object and hand detection quality on the validation set of Something-Something and EPIC-Kitchens using a TSM backbone on RGB.}
\label{tab:bbox_inf} 
\end{table}

\begin{table*}[!htb]
\centering
\resizebox{\linewidth}{!}{
\setlength{\tabcolsep}{7.3pt}
\begin{tabular}{@{}clccc ccc ccc ccc@{}}
\toprule
 & & \multicolumn{6}{c}{Overall} & \multicolumn{3}{c}{Unseen Participants} & \multicolumn{3}{c}{Tail Classes} \\
 \cmidrule(r){3-8} \cmidrule(lr){9-11} \cmidrule(l){12-14}
 & & \multicolumn{3}{c}{Top-1 Accuracy (\%)} & \multicolumn{3}{c}{Top-5 Accuracy (\%)} & \multicolumn{3}{c}{Top-1 Accuracy (\%)} & \multicolumn{3}{c}{Top-1 Accuracy (\%)} \\
 \cmidrule(r){3-5} \cmidrule(lr){6-8} \cmidrule(lr){9-11} \cmidrule(l){12-14}
Set & Baseline & \multicolumn{1}{c}{Verb} & \multicolumn{1}{c}{Noun} & \multicolumn{1}{c}{Act.} & \multicolumn{1}{c}{Verb} & \multicolumn{1}{c}{Noun} & \multicolumn{1}{c}{Act.}& \multicolumn{1}{c}{Verb} & \multicolumn{1}{c}{Noun} & \multicolumn{1}{c}{Act.} & \multicolumn{1}{c}{Verb} & \multicolumn{1}{c}{Noun} & \multicolumn{1}{c}{Act.} \\ \midrule
\multirow{7}{*}{\rotatebox{90}{\textbf{Val}}}
& TRN~\citep{zhou2018temporal} & 65.88 & 45.43 & 35.34 & 90.42 & 71.88 & 56.74 & 55.96 & 37.75 & 27.70 & 34.66 & 17.58 & 14.07 \\ 
& SlowFast~\citep{feichtenhofer2019slowfast} & 65.56 & \textbf{50.02} & 38.54 & 90.00 & \textbf{75.62} & 58.60 & 56.43 & \textbf{41.50} & 29.67 & 36.19 & 23.26 & \textbf{18.81} \\
& TSM~\citep{lin2019tsm} & 67.86 & 49.01 & 38.27 & 90.98 & 74.97 & \textbf{60.41} & 58.69 & 39.62 & 29.48 & 36.59 & \textbf{23.37} & 17.62 \\
& TROI & \textbf{68.80} & 49.22 & \textbf{38.90} & \textbf{91.18} & 72.46 & 58.86 & \textbf{60.85} & {39.53} & \textbf{30.05} & \textbf{37.61} & 21.58 & 17.84 \\\cmidrule{2-14}
& TSM + TSM & {68.24} & {50.03} & {39.06} & \underline{91.18} & {75.80} & \underline{61.67} & 58.59 & {40.84} & {29.67} & 36.02 & \underline{22.73} & 17.52 \\
& TSM + TROI & \underline{68.98} & \underline{50.10} & \underline{39.59} & 91.09 & \underline{75.94} & {61.64} & \underline{59.91} & \underline{41.03} & \underline{30.70} & \underline{36.99} & 22.42 & \underline{17.87} \\\cmidrule{2-14} 
& ViViT~\citep{arnab2021vivit} & 66.40 & 56.80 & 44.00 & - & - & - & - & - & - & -& - \\ 
& ORViT~\citep{herzig2021object} & 68.40 & 58.70 & 45.70 & - & - & - & - & - & - & -& - \\ 
& MoViNet~\citep{kondratyuk2021movinets} & 72.20 & 57.30 & 47.70 & - & - & - & - & - & - & -& - \\ 
\midrule 
\multirow{6}{*}{\rotatebox{90}{\textbf{Test}}}
& TRN~\citep{zhou2018temporal} & 63.28 & 46.16 & 35.28 & 88.33 & 72.32 & 55.26 & 57.54 & 41.36 & 29.68 & 28.17 & 13.98 & 12.18\\
& SlowFast~\citep{feichtenhofer2019slowfast} & 63.79 & \textbf{48.55} & 36.81 & 88.84 & \textbf{74.49} & 56.39 & 57.66 & \textbf{42.55} & 29.27 & 29.65 & \textbf{17.11} & \textbf{13.45}\\
& TSM~\citep{lin2019tsm} & 65.32 & 47.80 & \textbf{37.39} & 89.16 & 73.95 & \textbf{57.89} & \textbf{59.68} & 42.51 & \textbf{30.61} & 30.03 & 16.96 & \textbf{13.45}\\
& TROI & \textbf{65.98} & 47.19 & 37.33 & \textbf{89.58} & 72.67 & 57.29 & 59.42 & 41.09 & 30.02 & \textbf{30.19} & 15.06 & 13.31 \\\cmidrule{2-14}
& TSM + TSM & {65.46} & {48.31} & {37.64} & {89.54} & 73.78 & {58.19} & {58.76} & 42.48 & {30.39} & 29.41 & \underline{15.51} & 12.74 \\
& TSM + TROI & \underline{66.63} & \underline{48.98} & \underline{38.59} & \underline{89.94} & \underline{73.84} & \underline{58.62} & \underline{60.56} & \underline{43.58} & \underline{31.63} & \underline{29.80} & 15.02 & \underline{12.97} \\
\bottomrule
\end{tabular}}
\caption{EPIC-Kitchens validation and test server results. All methods use RGB and optical flow images. The best non-ensembled results are indicated in \textbf{bold}; best ensembled results are \underline{underlined}. The results of TRN, SlowFast and TSM are computed by~\cite{damen2020rescaling}. 
} 
\label{tab:ar_epic_results}
\end{table*}

\begin{table}[!htb]
\centering
\resizebox{\linewidth}{!}{
\setlength{\tabcolsep}{2.0pt}
\begin{tabular}{@{}lccccc@{}}
\toprule
Method & \# frames & \multicolumn{2}{c}{1-Clip} & \multicolumn{2}{c}{2-Clip} \\ \midrule
 & & Top-1 & Top-5 & Top-1 & Top-5 \\\midrule
STIN~\citep{materzynska2020something} (GT) & - & - & - & 60.2 & 84.4\\
TROI (GT) & 8 & 62.6 & 87.7 & 64.9 & 89.3 \\
TROI (GT) + scene & 8 & 64.4 & \textbf{88.8} & 66.4 & \textbf{90.2} \\
TROI (GT) + scene + coord & 8 & \textbf{64.7} & 88.6 & \textbf{66.8} & 90.1 \\\midrule
STIN~\citep{materzynska2020something} & - & 37.2 & 62.4 & - & -\\ 
STRG~\citep{wang2018videos} & - & 52.3 & 78.3 & - & -\\
STIN + STRG & - & 56.2 & 81.3 & - & -\\ 
Non-Local I3D~\citep{wang2018non}& 32 &44.4 & 76.0 & - & -\\ 
TRN~\citep{zhou2018temporal} & 8 & 48.8 & 77.6 & -& -\\
TSM~\citep{lin2019tsm} & 8 & 58.8 & 85.6 & 61.3 & 87.3 \\
RubiksNet~\citep{fanbuch2020rubiks} & 8 & 59.0 & 85.2 & 61.7 & 87.3 \\
bLVNet-TAM~\citep{fan2019more} & $8\times2$\footref{fn1}& 59.1 & \textbf{86.0} & -& -\\
TROI (Pred.) + scene & 8 & \textbf{59.4} & 85.6 & \textbf{62.0} & \textbf{87.4} \\\midrule
TSM~\citep{lin2019tsm} & 16 & 61.4 & 87.0 & 63.4 & 88.5 \\
bLVNet-TAM~\citep{fan2019more} & $16\times2$\footref{fn1}& 61.7 & 88.1 & -& -\\
ViViT~\citep{arnab2021vivit} & 16 & - & - & 65.9 & 89.9 \\
TDN~\citep{wang2021tdn}& 16 & - & - & \textbf{ 66.9 } & \textbf{ 90.9 }\\ 
CT-Net~\citep{li2020ct} & 16 & - & - & 64.5 & 89.3 \\ 
MVFNet \citep{wu2021mvfnet} & 16 & - & - & 65.2 & - \\ 
ORViT~\citep{herzig2021object} & 32 & - & - & \textbf{ 67.9 } & \textbf{ 90.5 }\\ 
MoViNet~\citep{kondratyuk2021movinets}& 50 & - & - & 64.1 & 88.8 \\ 
TROI (GT)& 16 & \textbf{65.0} & \textbf{89.2} & \textbf{66.9} & \textbf{90.2} \\ \bottomrule
\end{tabular}} 
\caption{Comparisons with SOTA architectures on Something-Something's validation set. Models are trained on RGB input and built on a Resnet-50 backbone using different number of frames as input. Reported are Top-1\&-5 accuracies over 1-\&-2 Clip settings. 
}
\label{tab:something_v2} 
\end{table}

\begin{table}[!htb] 
\centering
\resizebox{\linewidth}{!}{
\setlength{\tabcolsep}{35.1pt}
\begin{tabular}{@{}lll@{}}
\toprule
IOU @0.50 & IOU @0.75 & IOU @0.90 \\ \midrule
79.21 & 59.58 & 18.56 \\ \bottomrule
\end{tabular}}
\caption{IOUs of $\alpha$ = \{50\%, 75\%, 90\%\} computed between the tracking predictions~\citep{bewley2016simple,wu2019detectron2} and the GT bounding boxes from ~\cite{materzynska2020something} on the Something-Something dataset.}
\label{fig:ioudets}
\end{table}

\begin{figure}[t]
\centering
\includegraphics[width=.45\linewidth]{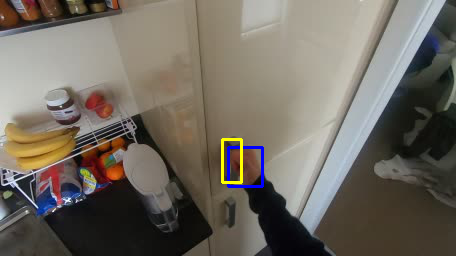} 
\includegraphics[width=.45\linewidth]{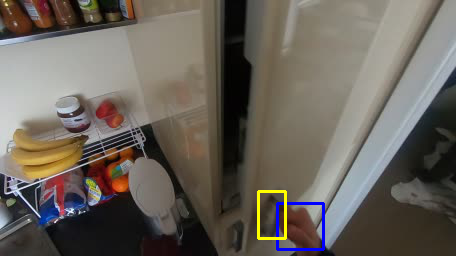}\\ \vspace{1.2mm}
\includegraphics[width=.45\linewidth]{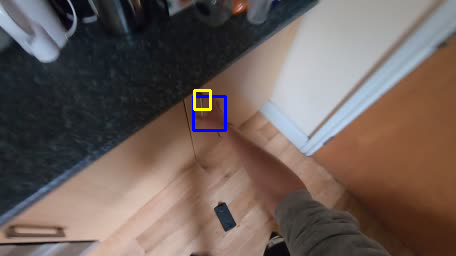} 
\includegraphics[width=.45\linewidth]{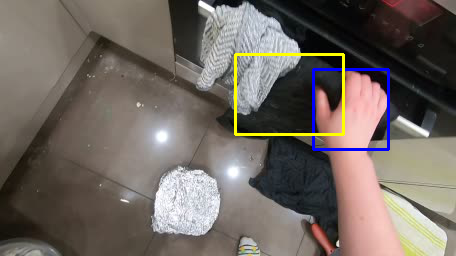} 
\caption{Qualitative examples of misleading object localizations in \emph{``open fridge''} (top row), \emph{``open cupboard''} (bottom left) and \emph{``open oven''} (bottom right).}
\label{fig:hand_obj_wrong}
\end{figure}

\begin{table}[!htb] 
\centering
\resizebox{\columnwidth}{!}{
\setlength{\tabcolsep}{10.1pt}
\begin{tabular} {lll}\toprule
Noun & Imp.(\%) & Detection \textcolor{white}{Detection Detection Detection.} \\ \hline 
 mushroom & +12.50\% & \parbox[c]{1em}{\includegraphics[height=1.19in]{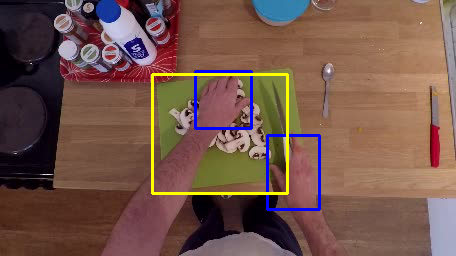}} \\ \hline 
 board & +9.40\% & \parbox[c]{1em}{\includegraphics[height=1.195in]{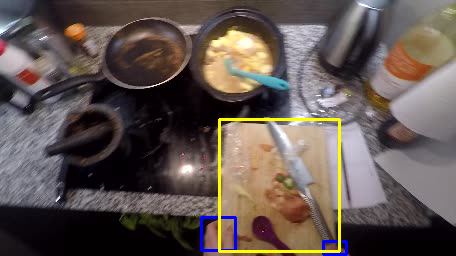}} \\ \hline 
 bowl & +4.51\% & \parbox[c]{1em}{\includegraphics[height=1.195in]{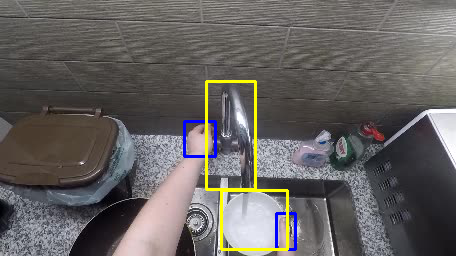}} \\ \hline 
 pan & +4.42\% & \parbox[c]{1em}{\includegraphics[height=1.195in]{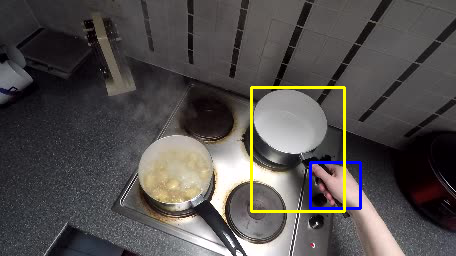}} \\ \hline 
 lid & +2.90\% & \parbox[c]{1em}{\includegraphics[height=1.195in]{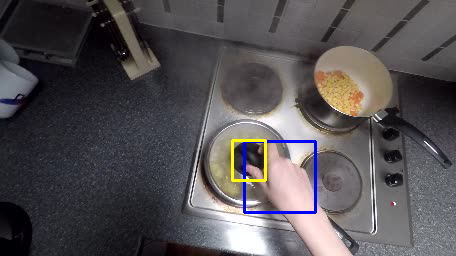}} \\ \hline 
 spatula & +1.64\% & \parbox[c]{1em}{\includegraphics[height=1.195in]{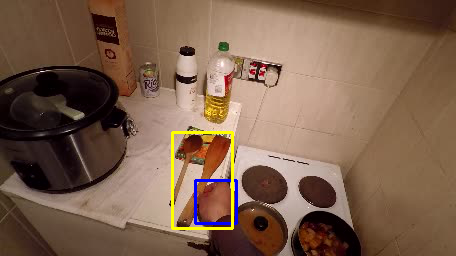}} \\ \hline 
\bottomrule
\end{tabular}}
\caption{
Qualitative examples of improved (``Imp.'') noun classes over TSM on EPIC-Kitchens.
Percentages are differences in accuracy per noun class.
We can see that the hand and objects localizations are accurate for these nouns. 
} 
\label{tab:positive1}
\end{table}

\begin{table}[!htb] 
\centering
\resizebox{\columnwidth}{!}{
\setlength{\tabcolsep}{10.1pt}
\begin{tabular} {lll}\toprule
Noun & Deg.(\%) & Detection \textcolor{white}{Detection Detection Detection.} \\ \hline 
dishwasher & -11.11\% & \parbox[c]{1em}{\includegraphics[height=1.15in]{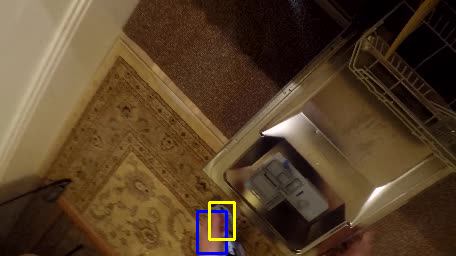}} \\ \hline 
scissors & -10.14\% & \parbox[c]{1em}{\includegraphics[height=1.2in]{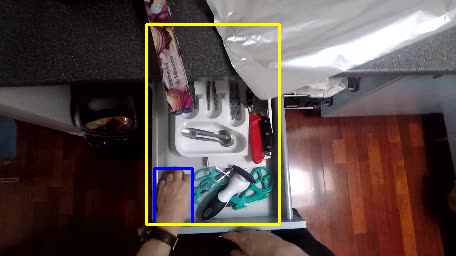}} \\ \hline 
microwave & -5.88\% & \parbox[c]{1em}{\includegraphics[height=1.2in]{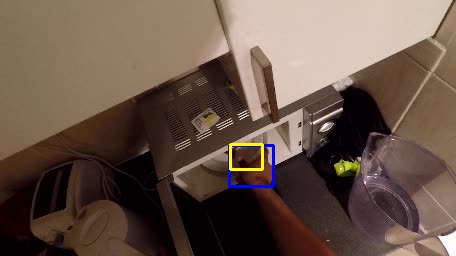}} \\\hline 
oven & -5.21\% & \parbox[c]{1em}{\includegraphics[height=1.2in]{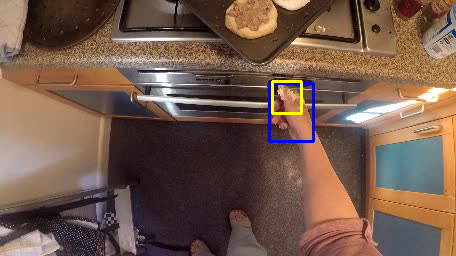}} \\ \hline 
cupboard & -1.92\% & \parbox[c]{1em}{\includegraphics[height=1.2in]{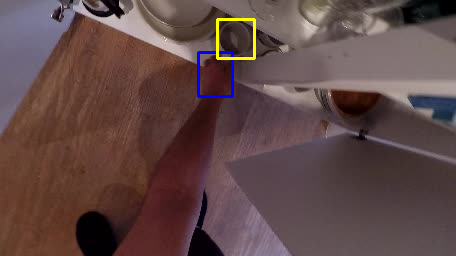}} \\ \hline 
fridge & -1.32\% & \parbox[c]{1em}{\includegraphics[height=1.2in]{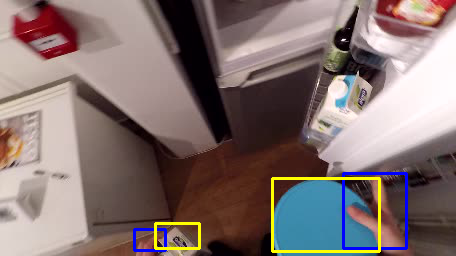}} \\ \hline 
\bottomrule
\end{tabular}}
\caption{Qualitative examples of degraded (`` Deg.'') noun classes due to misleading object localizations on EPIC-Kitchens.
Percentages represent differences in accuracy per noun class compared to TSM.}
\label{tab:negative1}
\end{table}

\textbf{Detection Quality:} How much impact does the quality of the detections have during inference? To check this, we test our models with either missing or corrupted ROIs and show the results in Table~\ref{tab:bbox_inf}. Using \emph{``only hands''} regions decreases the performance more compared to using \emph{``only objects''} by 5.7\% for Something-Something and 2.4\% for EPIC-Kitchens. Selecting only the right hand and right object, \emph{``r hand \& obj''}, performs better than selecting the entities on the left side, \emph{``l hand \& obj''}, on EPIC-Kitchens, with a difference of 0.6\%. This is likely due to the imbalance of right-handed participants in this dataset. When no ROIs are used during inference, \ie our module is bypassed completely, our performance decreases by more than 8\% for Something-Something and 4\% for EPIC-Kitchens. This decrease is unsurprising as our model is trained with the module in place for all samples and the backbone is adapted to the transformed ROIs on the feature map after the conv4 layer.

Finally, we corrupt the ROIs by shifting the ROIs to decrease the IoU with the original bounding boxes, leaving an IoU of $\alpha$ = \{50\%, 25\% 5\%\}. This experiment corresponds to explicitly putting emphasis (attending) on wrong ROIs, \ie the background in the middle layers of the CNN. Our model simply transforms the wrong regions and replaces them on the overall feature map; this overall feature map, however, still contains visual cues of the object and hand. Our performance does not decrease significantly, even with 50\% corruption for both datasets. However, decreasing the overlap to 5\%, which corresponds to an intentional focus on the background regions, decreases our performance by 8\% on Something-Something and by 6\% on EPIC-Kitchens.

\begin{figure*}[!htb]
\centering 
\subfloat[Putting smth similar on table]{\label{fig:mdleft}{\includegraphics[width=0.21\textwidth]{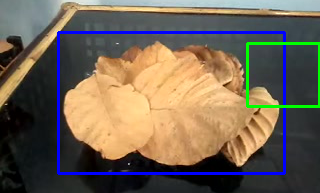}}}\hfill 
\subfloat[Throwing smth onto a surface]{\label{fig:mdleft}{\includegraphics[width=0.21\textwidth]{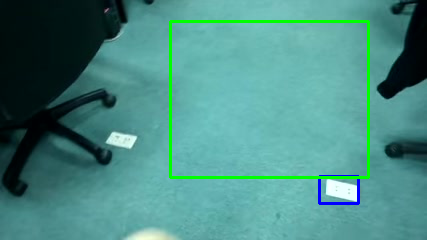}}}\hfill
\subfloat[Putting smth and smth on table]{\label{fig:mdleft}{\includegraphics[width=0.21\textwidth]{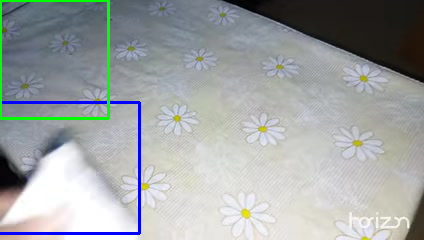}}}\hfill
\subfloat[Throwing smth onto a surface ]{\label{fig:mdleft}{\includegraphics[width=0.21\textwidth]{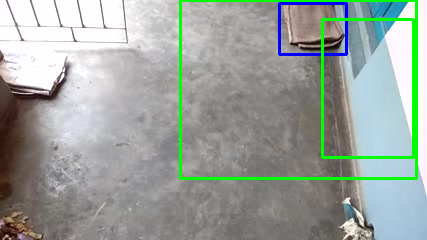}}}\hfill
\subfloat[Putting smth on a surface]{\label{fig:mdleft}{\includegraphics[width=0.21\textwidth]{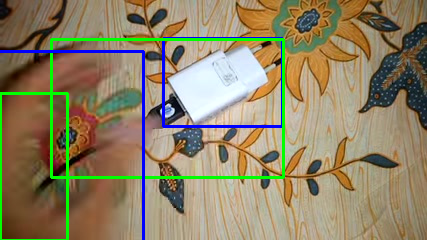}}}\hfill
\subfloat[Covering smth with smth ]{\label{fig:mdleft}{\includegraphics[width=0.21\textwidth]{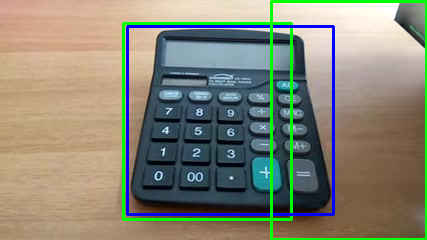}}}\hfill
\subfloat[Picking smth up ]{\label{fig:mdleft}{\includegraphics[width=0.21\textwidth]{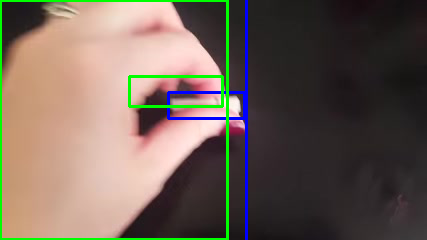}}}\hfill 
\subfloat[Putting smth on a surface]{\label{fig:mdleft}{\includegraphics[width=0.21\textwidth]{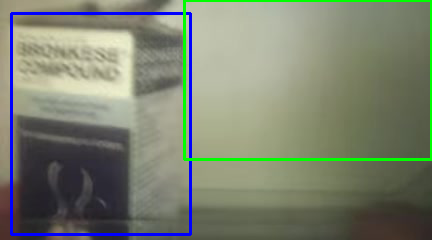}}}\hfill
\caption{
For Something-Something, we visualize tracking predictions from~\cite{bewley2016simple,wu2019detectron2} and GT bounding boxes. Blue corresponds to GT and green to predictions. 
We observe that the prediction outputs might include missing bounding boxes and bounding boxes with no or little overlap with the GT. 
}
\label{tab:predBad}
\end{figure*}

\subsection{Comparison to the State-of-the-Art (EPIC-Kitchens-100)}

We perform a comparison with state-of-the-art action recognition methods on EPIC-Kitchens in Table~\ref{tab:ar_epic_results}. Like TSM~\citep{lin2019tsm} and SlowFast~\citep{feichtenhofer2019slowfast}, we train two models on RGB and optical flow images individually to predict both verbs and nouns. During inference, we average the pre-softmax predictions from the two models. 
We note that because the validation and test sets are so large (9.7k / 13.1k instances, respectively), achieving a gain on the order of 0.1 can be quite challenging. 
In the EPIC-Kitchens-55 report~\citep{epicreport20}, the difference between top submissions is 0.97\% \& 0.22\% for seen and 0.58\% \& 0.41\% for unseen kitchens.

In Table~\ref{tab:ar_epic_results}, TROI with a TSM backbone outperforms TRN and SlowFast for overall and unseen action accuracies on both the validation and test sets. It achieves higher verb scores than SlowFast (+3.2\% and +2.2\% in overall Top-1 accuracy), but the noun scores degrade (-0.4\% and -1.4\%). On the validation set, we improve the action accuracy over TSM~\citep{lin2019tsm} on overall, unseen and classes by 0.6\%, 0.6\% and 0.2\%. However, on the test set, we decrease the unseen and tail action accuracy by 0.6\% and 0.1\%. Similar to our comparison to SlowFast, our verb accuracy outperforms TSM in almost all cases, but the noun accuracy is generally lower. 

TROI's poor performance on the noun accuracy, especially in the tail classes, is at first glance puzzling, since it is supplied with interacting object ROIs. We further break down the classes of EPIC-Kitchens and observe that we improve over TSM on actions involving small objects like spoons or knives and mid-sized objects like pans or bowls. But on large objects like fridges or cupboards we do worse. Visualizing the bounding boxes for these classes, we observe that the detection algorithm of \cite{shan2020understanding} has a strong size prior and localizes object parts such as the fridge or oven handle (see Figure~\ref{fig:hand_obj_wrong}). 
We speculate that in these cases, our TROI module does not get sufficient visual context from the interacting objects or simply gets wrong context, which decreases our model's accuracy.
Since EPIC-Kitchens-100 does not provide GT, we cannot evaluate bounding box accuracy. 
We, therefore, qualitatively check the classes of EPIC-Kitchens to examine where our model improves and where it degrades compared to TSM. In Tables~\ref{tab:positive1}, we show the classes where our TROI module improves the noun scores over TSM. These examples show that the hand and objects localizations are accurate. In Tables~\ref{tab:negative1}, we present the degraded noun classes with TROI. For each noun, we give the total decrease compared to TSM. We also present an example frame to show misleading localizations. 
Comparing Tables~\ref{tab:positive1} and ~\ref{tab:negative1}, we can say that the entire object or appliance needs to be localized to identify the noun and therefore the action. 

To compensate for this weakness, we test an ensemble (TSM + TROI), fusing our predictions with TROI with a standard TSM. The two complement each other well, as the ensemble provides a considerable gain over both our own work and TSM individually. It also edges out an ensemble of two separately trained TSMs (TSM + TSM). Additionally, we improve upon the state-of-the-art on the validation set, SlowFast, by 1.8\%, and the state-of-the-art on the test set, TSM, by 1.2\%.

Table~\ref{tab:ar_epic_results} also performs a comparison with more recent transformer-based architectures for the validation set of EPIC-Kitchens-100. Note that these networks are heavier parameter-wise and also process more than 8 frames at once. Our model is competitive on the verb scores, but our noun accuracy is generally low, highlighting the importance of improving the performance of the object detector and the superiority of transformer-based backbones. 
 
\subsection{Comparison to the State-of-the-Art (Something-Something)}

We perform a comparison to the state-of-the-art in Table~\ref{tab:something_v2} for the validation set for the 1- and 2-Clip settings with 8 or 16 frames as input. To ensure a fair comparison, we use RGB images to match competing methods, although incorporating flow would likely yield further improvements. Similar to our work is STIN, which also uses localized object and hands and models geometric configurations between the detections. Without requiring an ensemble with a separately trained appearance model, our unified framework (TROI (GT)), outperforms ``STIN (GT)'' by more than 4.6\%. As a variant, we add additional information in the form of max-pooled features from the entire frame (TROI (GT) + scene) and then add the bounding box coordinates into the positional encoding (+ coord). The additional information gives further gains of about 2\%. We note that these additional variants do not have much impact for EPIC; this is likely due to the cluttered scenes in EPIC-Kitchens in contrast to Something-Something. 

Our approach, TROI with a TSM backbone, outperforms all models for different inputs when trained and tested with prediction outputs. In particular, the gap is quite significant for TRN and STIN, with 10.6\% and 22.1\% differences in Top-1 accuracies. Compared to the TSM baseline, we have smaller gains (+0.6\% in 1-Clip vs. +0.7\% in 2-Clip). Our module also outperforms the two state-of-the-art methods bLVNet-TAM~\citep{fan2019more}\footnote{\label{fn1}Two-branch architecture with 2 input streams of 8/16 frames each.} and RubiksNet~\citep{fanbuch2020rubiks}. For ``TROI (Pred.)'', we train and test our model with ~\cite{materzynska2020something}'s tracking outputs. Compared to using GT boxes, our performance decreases by about 5\%. We speculate that the drop is due to the quality of the tracking results. 
We further investigate the performance of the tracking outputs. In Table~\ref{fig:ioudets}, we present the IOUs of overlap thresholds, $\alpha$ = \{50\%, 75\%, 90\%\}, computed between the tracking predictions and the GT bounding boxes. The IOU decreases significantly when $\alpha$ increases. We break down the classes of Something-Something and visualize some predicted vs. GT bounding boxes in Figure~\ref{tab:predBad}. We observe that the prediction outputs include missing bounding boxes and bounding boxes with no or little overlap with the GT. Moreover, training with GT bounding boxes and using predicted boxes during inference reduces the Top-1 \& 1-Clip performance by 1.4\%, indicating a domain gap between the two. Nevertheless, the accuracy drop also indicates that there is still room for improving robustness to poorly localized ROIs. One possibility is to reduce the gap by leveraging both ground truth bounding boxes and estimated ROIs simultaneously and incorporate a self-supervised loss to predict the box's source. We leave such an extension for future work.

With 16 frames as input, we achieve even stronger performance and improve the accuracy of 8 frames by 2.3\% and 2\% for the 1- and 2-Clip settings, respectively. However, this comes at the expense of more computation. We outperform bLVNet-TAM~\citep{fan2019more} and TSM by more than 4\%. 
We also compare against more recent transformer-based architectures, ViViT~\citep{arnab2021vivit}, TDN~\citep{wang2021tdn}, CT-Net~\citep{li2020ct}, MVFNet \citep{wu2021mvfnet}, ORViT~\citep{herzig2021object}, and MoViNet~\citep{kondratyuk2021movinets}. We outperform these models or are comparable to the methods using 16 frames. A very recent work, ORViT~\citep{herzig2021object}, using 32 frames, outperforms all works. Our model's competitive performance demonstrates the effectiveness of augmenting convolutional networks with spatio-temporal relations and exploring transformer-based backbones.
 
\section{Conclusion \& Outlook}
The relations and transformations of human-centered entities such as hands or interacting objects are critical cues for video recognition. We here introduce a relation module that can be integrated into standard CNNs to model both short and long-range interactions among such entities. Based on these interactions, we transform localized regions of interest within the feature map via self-attention. Our model is effective on two large-scale action recognition datasets, Something-Something and EPIC-Kitchens-100. Our framework's gains over the state-of-the-art highlight the importance of integrating temporal and relational information for action recognition from videos. It also opens up a new avenue of exploration for relational models via mid-level features. In the future, we will extend our module by making it more robust to missing detections and wrongly identified regions. \\

\noindent\textbf{Declaration of competing interest}\\
The authors declare that they have no known competing financial interests or personal relationships that could have appeared to influence the work reported in this paper.\\

\noindent\textbf{Acknowledgments}\\
This research / project is supported by the National Research Foundation, Singapore, under its NRF Fellowship for AI (NRF-NRFFAI1-2019-0001).

\bibliographystyle{model2-names}
\bibliography{refs}

\end{document}